\documentclass[letterpaper]{article} 
\usepackage[submission]{aaai2026}  
\usepackage{times}  
\usepackage{helvet}  
\usepackage{courier}  
\usepackage[hyphens]{url}  
\usepackage{graphicx} 
\usepackage{natbib}  
\usepackage{caption} 
\usepackage{xspace}
\usepackage{amssymb} 
\usepackage{amsmath}

\usepackage{bm}
\usepackage{booktabs}
\usepackage{multirow}
\usepackage{makecell}
\usepackage[linesnumbered,ruled,vlined]{algorithm2e}
\usepackage{algpseudocode}
\usepackage{soul}
\usepackage{footnote}
\usepackage{diagbox}
\usepackage{supertabular}
\usepackage{tabularx}

\usepackage{array}
\usepackage{siunitx} 
\usepackage{listings} 
\usepackage{enumitem}

\newcommand{\ie}{\emph{i.e.,}\xspace}
\newcommand{\structpbo}{\emph{StructPBO}\xspace}
\newcommand{\autopbo}{\emph{AutoPBO}\xspace}
\newcommand{\nupbo}{\emph{NuPBO}\xspace}
\newcommand{\orasls}{\emph{OraSLS}\xspace}
\newcommand{\gurobi}{\emph{Gurobi}\xspace}
\newcommand{\scip}{\emph{SCIP}\xspace}
\newcommand{\pboihs}{\emph{PBO-IHS}\xspace}
\newcommand{\roundingsat}{\emph{RoundingSat}\xspace}
\newcommand{\role}{\emph{Role}\xspace}
\newcommand{\tasks}{\emph{Tasks}\xspace}
\newcommand{\tips}{\emph{Tips}\xspace}
\newcommand{\win}{$\#win$\xspace}
\newcommand{\avgscore}{$avg\_{score}$\xspace}
\newcommand{\minisat}{\emph{MINISAT+}\xspace}
\newcommand{\openwbo}{\emph{Open-WBO}\xspace}
\newcommand{\naps}{\emph{NaPS}\xspace}
\newcommand{\satj}{\emph{sat4j}\xspace}
\newcommand{\lspbo}{\emph{LS-PBO}\xspace}
\newcommand{\decilspbo}{\emph{DeciLS-PBO}\xspace}
\newcommand{\dlspbo}{\emph{DLS-PBO}\xspace}

\newcommand{\MC}[1]{#1}

\newcommand{\newMC}[1]{#1}

\usepackage{newfloat}
\usepackage{listings}
\DeclareCaptionStyle{ruled}{labelfont=normalfont,labelsep=colon,strut=off} 
\title{AutoPBO: LLM-powered Optimization for Local Search PBO Solvers}

\author{
    Yi Chu\textsuperscript{\rm 3},  
    Jinyuan Li\textsuperscript{\rm 1, \rm 2},
    Yiwen Sun\textsuperscript{\rm 4},
    Mengchuan Zou\textsuperscript{\rm 1},
    Shaowei Cai\textsuperscript{\rm 1, \rm 2}\footnote{Corresponding author}
}

\affiliations{
    \textsuperscript{\rm 1}Key Laboratory of System Software, Institute of Software, Chinese Academy of Sciences, Beijing, China\\
    \textsuperscript{\rm 2}School of Computer Science and Technology, University of Chinese Academy of Sciences, Beijing, China\\
    \textsuperscript{\rm 3}Institute of Software, Chinese Academy of Sciences, Beijing, China\\
    \textsuperscript{\rm 4}School of Data Science, Fudan University, Shanghai, China\\
    chuyi2020@iscas.ac.cn, lijy@ios.ac.cn, ywsun22@m.fudan.edu.cn, zoumc@ios.ac.cn, caisw@ios.ac.cn
}

\usepackage{bibentry}

\begin{document}

\maketitle

\begin{abstract}
Pseudo-Boolean Optimization (PBO) provides a powerful framework for modeling combinatorial problems through pseudo-Boolean (PB) constraints. Local search solvers have shown excellent performance in PBO solving, and their efficiency is highly dependent on their internal heuristics to guide the search. Still, their design often requires significant expert effort and manual tuning in practice. While Large Language Models (LLMs) have demonstrated potential in automating algorithm design, their application to optimizing PBO solvers remains unexplored. In this work, we introduce \autopbo, a novel LLM-powered framework to automatically enhance PBO local search solvers. 
We conduct experiments on a broad range of four public benchmarks, including one real-world benchmark, a benchmark from PB competition, an integer linear programming optimization benchmark, and a crafted combinatorial benchmark, 
to evaluate the performance improvement achieved by \autopbo and compare it with six state-of-the-art competitors, including two local search PBO solvers \nupbo and \orasls, two complete PB solvers \pboihs and \roundingsat, and two mixed integer programming (MIP) solvers \gurobi and \scip. 
\autopbo demonstrates significant improvements over previous local search approaches, while maintaining competitive performance compared to state-of-the-art competitors. The results suggest that \autopbo offers a promising approach to automating local search solver design.

\end{abstract}


\section{Introduction}
\label{sec:Introduction}
Pseudo-Boolean Optimization (PBO) plays \MC{an important role in solving a wide range of combinatorial problems \cite{boros2002pseudo}, which seeks an assignment of values to a set of Boolean variables that optimizes a linear objective function under Pseudo-Boolean (PB) constraints}. Due to its powerful expressiveness and \MC{the convenience to make use of properties of boolean variables}, PBO has demonstrated broad applicability across \MC{various} domains, including VLSI design, economic modeling, computer vision, and manufacturing optimization \cite{Wille_Zhang_Drechsler_2011, Zhang_Hartley_Mashford_Burn_2011, roussel2021pseudo}.

\MC{\newMC{Along with} the wide usages of the PBO problem in industrial and application domains, solving the PBO is then a non-negligible topic. However, the solving} of PBO is a challenging task as the problem is NP-hard \cite{buss2021proof}. \MC{In previous studies,} the solving methods can be divided into two classes: complete methods and incomplete methods. \MC{The complete methods solves the problem to optimal and proves the optimality, while incomplete methods do not guarantee to compute the optimal assignment but try to compute good solutions in a short time.}

Research on complete methods for solving PBO has developed multiple approaches. 
First, since PB constraints can be naturally treated as 0-1 linear constraints, mixed-integer programming (MIP) solvers such as \scip \cite{bestuzheva2021scip} and \gurobi \cite{gurobi2021} can be directly applied to solve PBO problems. 
Second, by translating PB constraints into conjunctive normal form (CNF), the problem can be solved using SAT solvers based on Conflict-Driven Clause Learning (CDCL), including \minisat \cite{een2006translating}, \openwbo \cite{martins2014open}, and \naps \cite{DBLP:journals/ieicet/SakaiN15}.
Beyond these, advanced methods have been developed for more efficient PBO solving. The cutting planes technique, which goes beyond the resolution power of CDCL, is implemented in solvers like \satj \cite{le2010sat4j} and \roundingsat \cite{roundingsat_v1,roundingsat_card,roundingsat_hybrid}. Additionally, the implicit hitting set (IHS) method has been successfully adapted to PBO in solvers such as \pboihs \cite{pboihs_2021,pboihs_2022}.


Complete algorithms often struggle with large-scale instances, leading to the development of incomplete approaches, among which local search stands out as a representative strategy \cite{lei2021efficient,chu2023towards,zhou2023improving}. 
The first notable solver \lspbo \cite{lei2021efficient} introduced a weighting scheme and a scoring function that jointly handle hard and soft constraints. 
Several key extensions followed: \decilspbo \cite{jiang2023decils} enhanced the framework by incorporating unit propagation; \nupbo \cite{chu2023towards} proposed enhanced scoring functions and weighting schemes; and \dlspbo \cite{chen2024parls} implemented dynamic scoring functions. 
Additionally, \orasls \cite{iser2023oracle} utilizes a oracle mechanism to guide the local search approach in PBO.

The efficiency of local search solvers heavily relies on internal heuristics to guide the search process. In the past, many works on designing different algorithmic components such as weighting scheme and score functions have been proposed, as in \cite{thornton2005clause,cai2013local,cai2014tailoring,cai2020old,lei2021efficient,chu2023aaai,ChuLY024}. However, those works are based on human-designed techniques and designing these heuristics often demands substantial expert effort and manual tuning in practice. On the other hand, recent developments on the LLM-based algorithm design start a new paradigm of algorithm design and show the capability of large language models (LLMs) in automating algorithm design. However, the application of LLMs to building PBO solvers remains unexplored, presenting a promising direction for future research. 

Current works on automated algorithm design for combinatorial optimization problems are mainly on designing evolutionary algorithms for specific problems or optimizing heuristics in simple solvers. FunSearch \cite{funsearch} pioneered the integration of pretrained LLMs with evolutionary search, initiating heuristic discovery through iterative code generation. Then, EoH \cite{eoh} extends this paradigm through dual-representation evolution, and ReEvo \cite{reevo} introduces a structured reflection mechanism to guide evolutionary search. What is more, AutoSAT \cite{autosat, automodsat} focus on optimizing heuristics in SAT solvers, AlphaEvolve \cite{alphaevolve} pushes the paradigm further by ensembling LLMs with automated evaluators in an evolutionary loop, enabling the discovery of entire algorithmic codebases.

\newMC{The works on automated heuristic design for the general form problem (i.e. problems with general constraints types, such as Integer Programming, Pseudo Boolean Optimization, etc) is rare, and current approaches of designing heuristics for specific types of problems face critical limitations in this scenario: 1) the general-problem solver usually have complex structure with various algorithm components, and thus results in long-context of source codes, being much more sophisticated than evolutionary algorithms for specific types of problems; 2) A general form problem usually allows a wide range of types of constraints rather than specific types of problems normally has quite limited and known types of constraints, making the heuristics design for these two scenarios quite different. Thus, the algorithm design for the general form optimization problem is still challenging, and to our knowledge, there is no prior work on the LLM-driven automated design for PBO solver.} 

\MC{We consider the automated optimization for local search solvers of the PBO problem. Specifically, we consider to enhance the existing state-of-the-art local search solver for PBO by leveraging the power of LLM. We try to address the above challenges by designing methods from three considerations:}1) Enhancing LLMs' comprehension of solver codes with complex structures; 2) Reducing errors or invalid modifications during code generation; 3) \newMC{Improving the solver's efficiency by optimizing its algorithm as a composition that involves multiple functions}.

In this work, we design a novel LLM-powered framework to automatically enhance PBO local search solvers. We propose a multi-agent system integrated with a greedy search strategy, enabling closed-loop, feedback-driven optimization. Furthermore, we introduce a structuralized local search PBO solver \structpbo, \newMC{with a clearer structure of codes, that could be used as an input for automatic optimization frameworks. This design helps LLMs to effectively comprehend and optimize PBO-specific search algorithms and was adopted in our system. }

 \newMC{Bring the above ideas together, we design our \autopbo framework to automatically enhancing local search solvers of PBO.} Experimental results demonstrate that \autopbo significantly improves the performance of local search PBO solvers, offering a promising approach to automating local search solver design.






\section{Preliminaries}
\label{sec:Preliminaries}
\subsection{Pseudo-Boolean Optimization}

A linear pseudo-Boolean (PB) constraint is expressed as:
$\sum_{j=1}^{n} a_j l_j \rhd b$, where $a_j,b \in \mathbb{Z}$ are integer coefficients, $b$ is the threshold, $\rhd \in \{=, >, \ge, <, \le\}$ is a relational operator, and each $l_j$ is a literal (either a Boolean variable $x_i$ or its negation $\neg x_i$).

In this work, we assume all PB constraints are in the normalized form $\sum_{j=1}^{n} a_j l_j \ge b \label{equa2}$ with $a_j, b \in \mathbb{N}_{0}^+$ (non-negative integers). 
This assumption is without loss of generality since all PB constraints can be converted to this form by expressing equalities as inequality pairs and applying the identity $x_i = 1 - \neg x_i$ to ensure non-negative coefficients \cite{roussel2021pseudo}.

An assignment $\alpha$ is a mapping from variables to $\{0,1\}$. 
A PB constraint $c$ is \emph{satisfied} under $\alpha$ if the inequality $\sum_{j=1}^{n} a_j l_j \ge b$ holds; otherwise, $c$ is \emph{violated}. 
The \emph{violation degree} of $c$ under $\alpha$, denoted $\operatorname{viol}(c)$, quantifies how far $c$ is from being satisfied:
$\operatorname{viol}(c) = \max\left(0, b - \sum_{j=1}^{n} a_j l_j \right)$,
which is zero if $c$ is satisfied, and otherwise measures the shortfall. 
A \emph{PB formula} $F$ is a conjunction of PB constraints, and an assignment that satisfies all constraints in $F$ is called a \emph{feasible solution}.

A pseudo-Boolean optimization (PBO) instance consists of a PB formula $F$ together with a linear Boolean objective function $\sum_{j=1}^n e_j l_j + d$, where $e_j \in \mathbb{N}^+$ and $d \in \mathbb{Z}$. Since all PB constraints must hold, they are treated as \emph{hard constraints}.  
For any assignment $\alpha$, its objective value is denoted as $obj(\alpha)$.  
A feasible solution $\alpha_1$ is considered superior to another solution $\alpha_2$ if $obj(\alpha_1) < obj(\alpha_2)$.  
The objective of PBO is to identify a feasible assignment $\alpha$ that minimizes $obj(\alpha)$.

\subsection{Local Search for PBO}
\label{sec:local search preliminary}
\MC{The local search is a general algorithmic paradigm fo solving combinatorial optimization problems.} It typically begins with an initial solution and iteratively explores the neighborhood of the current solution, seeking an improved candidate solution. If such a solution is found, it replaces the current one; otherwise, the search either terminates or employs strategies to escape local optima. The process continues until a stopping criterion is met, such as reaching a maximum number of iterations, a time limit, or a satisfactory solution quality threshold.

\MC{In a typical local search process of PBO,} given an instance $F$, the local search algorithm starts from an initial solution $\alpha$, then iteratively modifies $\alpha$ by selecting variables heuristically and applying corresponding \textbf{operators} (e.g., the $flip$ operator in pseudo-Boolean optimization) until a feasible solution is found. \MC{An operation is obtained when an operator is specified with a variable, and it is easy to see there could be multiple operations for generating a new solution. } During this process, \textbf{scoring functions} evaluate candidate operations, prioritizing operations that are likely to improve solution quality. An important factor normally included in scoring functions is the weights of constraints.
\MC{A \textbf{weighting scheme} is adopted to compute the weights of constraints in the scoring function, representing the importance of the constraints. }

In general, for greedy variable flipping, PBO local search (LS) algorithms employ a scoring function integrated with the weights of constraints. Let $w(c)$ denote the weight of a hard constraint $c$, and $w(o)$ denote the weight of the objective function $o$. For instance, in \lspbo, the penalty for a constraint $c$ is defined as $penalty(c) = w(c) \times viol(c)$, and the penalty for the objective function under the current assignment $\alpha$ is $penalty(o) = w(o) \times obj(\alpha)$. In \nupbo, a smoothed penalty method is proposed to balance the $viol$ values across constraints. Specifically, for a hard constraint $c$, the penalty function is redefined as:
$penalty(c)=\frac{ w(c)\times viol(c)}{smooth(c)}$
where $smooth(c)$ represents a smoothing coefficient derived from constraint properties. Similarly, for the objective function $o$, the penalty is adjusted to:
$penalty(o)=\frac{w(o)\times obj(\alpha)}{smooth(o)}$;
with $smooth(o)$ often calculated as the average of the objective function's coefficients. Across these algorithms, the hard score $hscore(x)$ of a variable $x$ quantifies the reduction in the total penalty of all hard constraints when $x$ is flipped. The soft score $oscore(x)$ measures the reduction in the objective function's penalty after flipping $x$. The scoring function of $x$ is defined as $score(x) = hscore(x) + oscore(x)$, a linear combination of the hard and soft scores.


\section{A New Structuralized Local Search PBO Solver: \structpbo}
\begin{algorithm}[t]
    {\caption{the structPBO solver}\label{alg:LS-PBO}}
    \KwIn{PBO instance $F$, cutoff time $cutoff$.}
    \KwOut{The best solution $\alpha^*$ found and its objective function value $obj^*$, or ``No solution found''.}

    $\alpha^* := \varnothing$, \quad $obj^* := +\infty$\;
    $\alpha := \text{InitializeAssignment}()$\; 
    \While{elapsed time $<$ cutoff}{
        \If{$\alpha$ is feasible \textbf{and} $obj(\alpha) < obj^*$}{

            \tcp*[h]{Update best solution and its objective function value}
            
            $\alpha^* := \alpha$, \quad $obj^* := obj(\alpha)$\;
        }
                
        \For{each variable $x$}{
            $\text{hscore}(x) := \Delta \text{Penalty}_{\text{hard}}(x)$\;
            $\text{oscore}(x) := \Delta \text{Penalty}_{\text{obj}}(x)$\;
            $\text{score}(x) := \text{CalculateScore}(\text{hscore}(x), \text{oscore}(x))$\;
        
        }
        
        \If{$D := \{x | \text{Score}(x) > 0\} \neq \varnothing$}{
        
            \tcp*[h]{A variable is picked accordingly}
            
            $x := \text{PickBestVariable}(D)$\;
        }
        \Else{
            \tcp*[h]{Stuck in a local optimum}
            
            UpdateWeights(F)\;

            \tcp*[h]{A variable is picked according to local-optima-escaping heuristics}
            
            $x := \text{PickEscapeVariable}(F)$\;
        }
        $\alpha := \alpha$ with $x$ flipped\;
    }
    
    \textbf{if} $\alpha^*\neq \varnothing$ \textbf{then} \Return $\alpha^*$ and $obj^*$\;
    \textbf{else} \Return No solution found\;
\end{algorithm}

\newMC{As we mentioned in Section \ref{sec:Introduction}, previous studies on automated algorithm design usually focus on combinatorial optimization problems with known types and usually result in simple-architecture algorithms, which are quite different from general form problem solvers that have complex structures. Notably, the Local Search PBO solvers typically employ advanced programming techniques and multiple algorithmic components in the code to achieve high efficiency in finding high-quality solutions within a short time period. Thus, it is still challenging for LLMs to process the codes of general problem solvers.   }

In our preliminary experiments, using LLMs to generate a PBO solver from scratch under the EoH mechanism resulted in local search algorithms that were limited to basic scoring functions and random perturbation stages (see Appendix A). Furthermore, when attempting to optimize existing Local Search PBO solvers using established solver optimization frameworks (see Appendix A), the high complexity and tight coupling of the codebase led to a significant increase in syntactic errors and logical inconsistencies in the LLM-generated code. 

\MC{To address these problems, we propose a predefined, structured Local Search PBO Solver framework named \structpbo}.
\MC{Aligning the basic components of local search routine, we defines the functionalities of different parts of a local search solver and build a structuralized solver \structpbo, following the SOTA PBO local search solver \nupbo\cite{chu2023towards}.}

As presented in Algorithm \ref{alg:LS-PBO}, a local search solver maintains a complete assignment (line 2). During the search, it keeps track of the best solution found (lines 4-5) and returns it when the termination condition is reached (lines 16-17). The effectiveness of the proposed local search PBO solver lies in its heuristic-driven variable selection mechanism, which hinges on the interplay between dynamic constraint weighting and penalty-based scoring functions, designed to systematically navigate the search space while balancing constraint satisfaction and objective optimization (lines 6-15). Specifically, for each variable $x$, the algorithm evaluates flip candidates through a composite score (line 9) that combines $hscore(x)$ and $oscore(x)$ – the respective penalty reductions for hard constraints and objective optimization (lines 7-8), where penalties are weighted according to constraint criticality. This weighting adapts dynamically: during feasibility-seeking phases, hard constraints dominate through exponentially higher weights, while objective constraints gain influence when approaching optimality. When trapped in local optima (line 13), the solver strategically updates these weights to escape stagnation. The resulting score-driven variable selection (lines 10-15) thus automatically shifts focus between constraint repair and objective improvement based on real-time search progress.

In this paper, we work on a structuralized local search framework that defines seven functions which are independently implemented: 
\begin{itemize}
    \item \textbf{InitializeAssignment}: Heuristically generates an initial complete assignment
    \item \textbf{Penalty\_hard}: Heuristically computes the hard constraint penalty reduction
    \item \textbf{Penalty\_obj}: Heuristically computes the objective penalty reduction
    \item \textbf{CalculateScore}: Heuristically combines hard and soft penalties into a dynamic composite score
    \item \textbf{PickBestVariable}: Heuristically selects the most promising variable
    \item \textbf{UpdateWeights}: Heuristically selects the most promising variable
    \item \textbf{PickEscapeVariable} : Heuristically identifies variables for diversification when stagnation is detected
\end{itemize}{}
This modular architecture enables isolated function optimization while maintaining global coherence through our convergence mechanism described in Section \ref{sec:greedy}.

\section{\MC{The Automated Optimization Framework}}
\label{sec:Main}

\begin{figure*}[ht!]
\centering
\includegraphics[width=\textwidth]{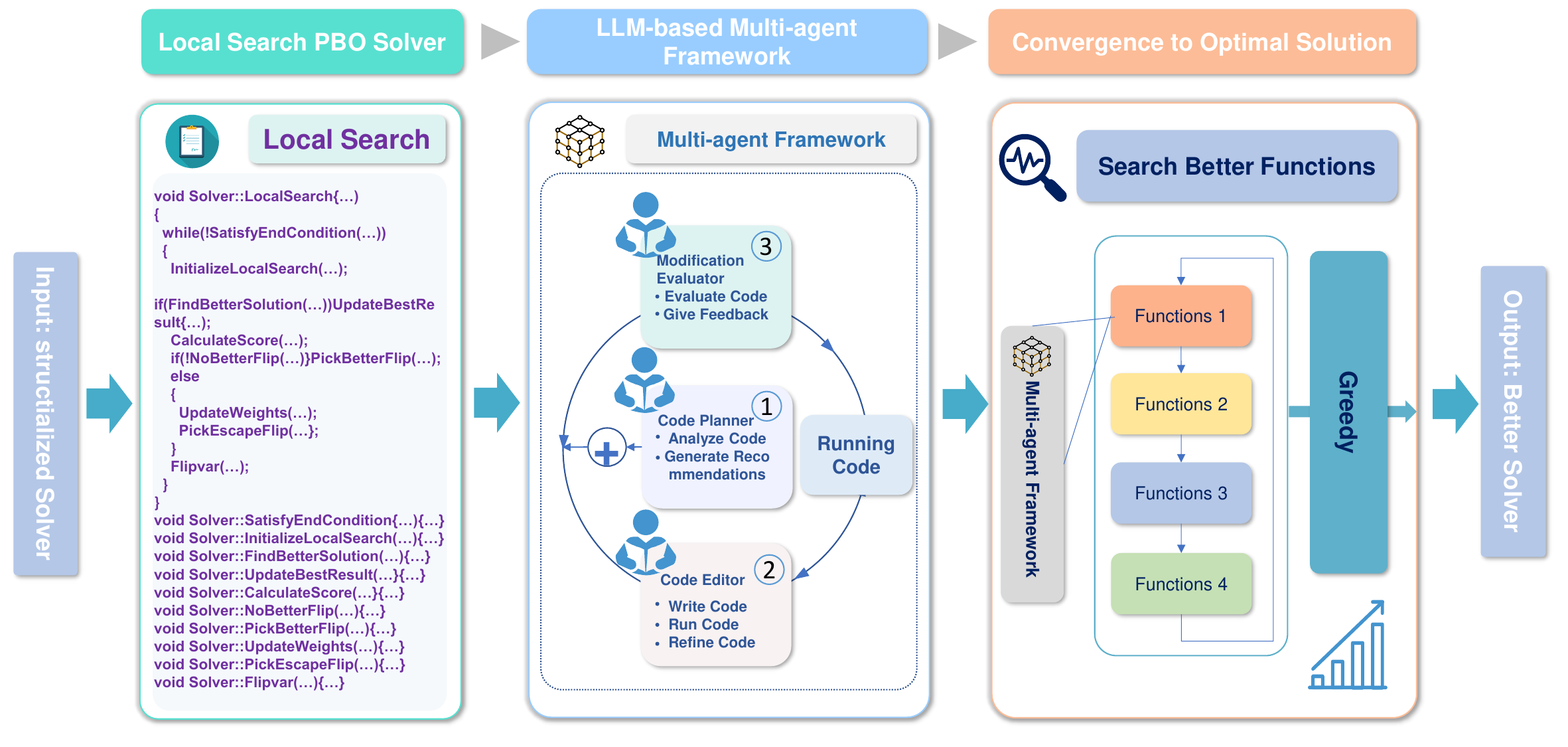} 
\caption{Architecture of AutoPBO. With a structuralized local search PBO solver as input, AutoPBO implements greedy strategy to optimize heuristic functions iteratively. 
    For each iteration, AutoPBO employs the original solver code to instantaneously create an understanding and recommendations for heuristic functions, subsequently engages in code generation and performance assessment by three agents. Upon completion, AutoPBO returns the optimal solver code. }
\label{fig:autopbo}
\end{figure*}

\MC{For automating the optimization of codes for local search solver of PBO, we propose a LLM-based multi-agent framework, named \autopbo. Our framework is based on three types of agents and a greedy-based iterative method to achieve a better performance solver. 
}

As illustrated in Figure \ref{fig:autopbo}, \MC{our framework} begins by loading a local search PBO solver, \MC{then a number of iterative code optimizing rounds are launched. In each round, we perform several distinct and independent code modifications work,  each modification work is realized by the three LLM agents collaboratively. After an optimization round, several versions of code are generated and then a greedy-based selection strategy is applied to select the most effective version for the next iteration. By repeating, the framework ultimately generates the resulting solver}.

\subsection{Optimization by Multiple Agents}

\begin{table*}[ht]
\centering
\includegraphics[width=\textwidth]{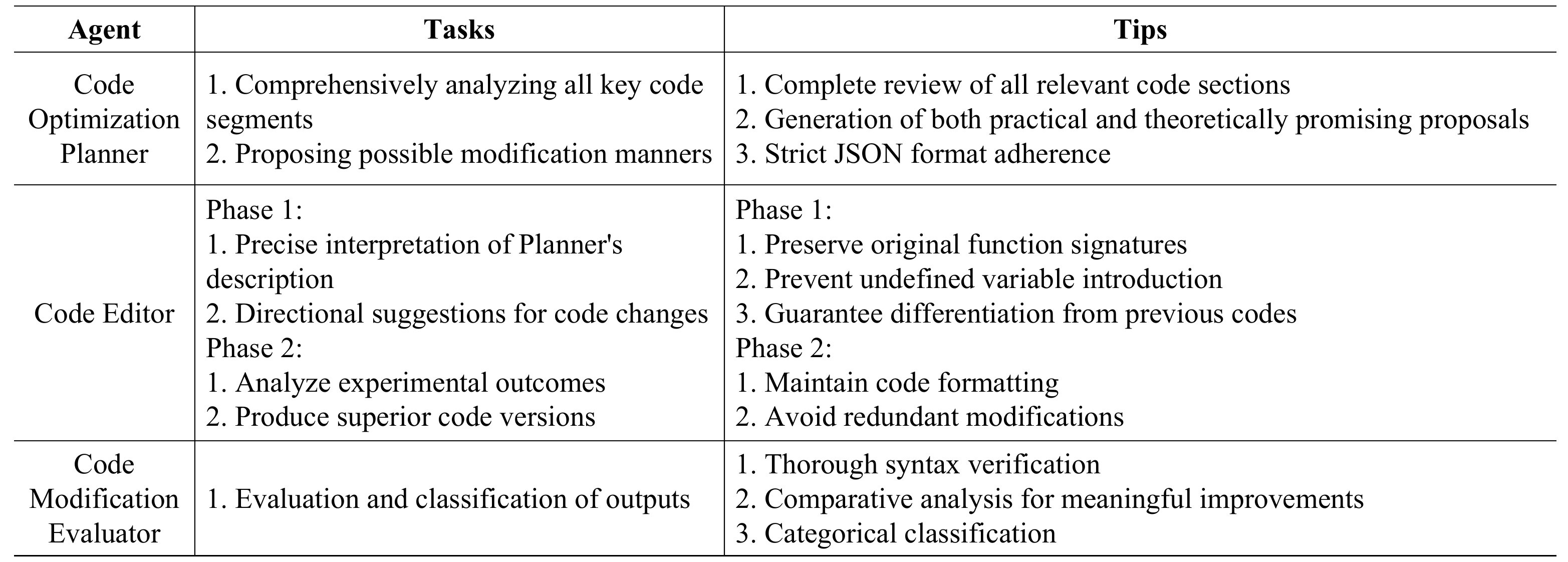} 
\caption{Agent Tasks and Tips Overview}
\label{tab:allprompt} 
\end{table*}
\label{sec:multi}
\MC{There are three specialized LLM agents in our framework, each of which has its own functionality and distinct roles, i.e., the Code Optimization Planner, Code Editor, and Modification Evaluator. Each of them serves different functionalities in our framework as follows:
\begin{itemize}
    \item Code Optimization Planner: Analyze key code segments to recognize the function targeted for modification and generate modification plans.
    \item Code Editor: Realize the code modification to improve the code, following the advice generated by the Code Optimization Planner.
    \item Modification Evaluator: Evaluate the modified code and generate advice for future improvement.
\end{itemize}
}

\MC{Those three agents operate through an automated cycle of modification plan generation, code edit iteration, and dynamic evaluation}.

\MC{A code optimization round is consist of two phases: the planning phase and the editing phase. In the planning phase, the Code Optimization Planner Agent identifies possible optimization ways and generates preliminary improvement plans, such as "Dynamic Score Ratios: Adjust h\_score\_ratio and s\_score\_ratio dynamically based on the current state of the search (e.g., increase the weight of hard constraints when the solution is infeasible'',  "Include a term that considers the age of the variable (time since last flip) to encourage diversification and escape local optima'', etc. The details of the implementations of the Code Optimization Planner is presented in Section 4.2.  }

\MC{The editing phase is built by interactions of the Code Editor and the Modification Evaluator, for multiple times. The Code Editor Agent performs the actual work of code modifications and runs for multiple times in the editing phase. Initially, it modifies the code modifications with the optimization plans provided by the Code Optimization Planner as input, and generates the first version of the code. It identifies if the modification has actual significant meanings, by excluding trivial modifications such as variable re-naming, parameter changing, etc. 
An actual compilation and running of the code is also performed at this time, to collect information such as the existence of compilation errors or running information. }

\MC{The running information and the evaluation results from the Modification Evaluator will then be sent to the code Editor as feedback for the next modification. This interaction will be performed by the Code Editor, Modification Evaluator several times, resulting in a final version of code for the current code optimization round. }

\MC{In the following, we will first present the implementation of different agents, then introduce the greedy-based iterative methods on top of code optimization rounds, which jointly form our whole optimization framework for PBO local search solver. }

\subsection{\MC{Implementation of Agents}}
The three agents are implemented by different prompt, which are designed following the principles proposed by OpenAI's foundational guidelines\cite{openai2023api}, we have developed a structured prompt to implement different functionalities for them.

Our prompt includes A \role set to the agent, the \tasks definitions, the \tips to provide suggestions and the complete PBO solver code appended at the end to ensure all agents share the same context. 
All three agents share a common \role configuration as a solver researcher attempting to improve the heuristics in a PBO solver. This foundational \role description establishes the professional identity and overarching objective for each agent in the optimization process. \tasks and \tips of every agent are shown in Table \ref{tab:allprompt}.

\subsection{Convergence to Optimal Solution}
\label{sec:greedy}
\MC{As illustrated in Figure 1, we implement an iterative greedy algorithm that progressively constructs the improved solver through three steps:}

\begin{itemize}
\item \textbf{Code Optimization Round}: Agents optimize one function at a time, while keeping others functions fixed. For example, we first generate multiple optimized versions of the UpdateWeights function through LLM agents, then select the best-performing version before proceeding to optimize CalculateScore. 
Implementation Process is that the LLM agents first generate multiple optimized versions of the function as illustrated in \ref{sec:multi}. Then our framework automates the following workflow: it constructs solver instances for each function version, runs these solvers in parallel on certain dataset, and automatically collects their outputs—including feasibility status (Feasible or Infeasible) and objective values($obj$) across all test instances. Finally, the framework determines the best-performing version through automatic comparison by tallying the total feasible solutions (Feasible count) and the number of instances where a solver's $obj$ outperforms \structpbo (Win count), selecting the version that achieves the highest count on both metrics.
\item \textbf{Modification Propagating}: The selected optimal function (e.g., improved UpdateWeights) is immediately integrated into \structpbo. Subsequent optimizations (e.g., CalculateScore refinement) are then performed on this updated \structpbo.
\item \textbf{Iterative Improvement}: This process repeats until all target functions are optimized.
\end{itemize}

This approach addresses the inherent dependency between functions. Consider the interaction between UpdateWeights and CalculateScore: The scoring function in CalculateScore must reflect the latest clause weights from UpdateWeights to properly prioritize constraint satisfaction. If we independently optimize both functions and combine their best versions, the scoring mechanism might ignore updated weight distributions, leading to inconsistent optimization behavior. Our greedy strategy prevents such conflicts by enforcing sequential adaptation - the improved CalculateScore automatically adapts to the newly integrated UpdateWeights through Modification Propagating.

\MC{Together with these three steps and the three agents involved, we iteratively improve our code, trying to get a better performance solver. }


\section{Experiments}




In this section, we introduce the experimental settings and present extensive experiments on 4 PBO benchmarks.
First, we evaluate the effectiveness of \autopbo as a framework for enhancing baseline solvers, demonstrating that it consistently improves performance across multiple benchmarks.
Second, we compare \autopbo with state-of-the-art PBO solvers to highlight its strong competitiveness.
Finally, in Appendix A we provide variance analysis through repeated experiments to demonstrate the statistical stability of \autopbo's performance.
\subsection{Settings}
\subsubsection{Environment} 
Our PBO solver is implemented in C++, while the interface for interacting with LLMs is developed in Python. The solver is compiled using g++ 9.4.0. All experiments are performed on an Ubuntu 20.04.4 LTS server, which is equipped with two AMD EPYC 7763 processors. Each processor operates at a base frequency of 2.45 GHz. The server is configured with 1TB of RAM. For all experiments, the DeepSeek\footnote{We utilize the DeepSeek-R1 as default in this paper.} LLM is employed. 

\subsubsection{Benchmarks and Datasets}
We evaluate \autopbo on four widely-used PBO benchmarks, comprising 47 datasets in total. For each dataset, we randomly split the instances into training and test sets with a 1:1 ratio. The training set is used to generate the enhanced solver via \autopbo, while the test set is reserved for final evaluation. Detailed instance information is provided in Code \& Data Appendix. A summary is provided below:
\begin{itemize}
    \item \texttt{PB16}: The OPT-SMALLINT-LIN benchmark from the 2016 pseudo-Boolean competition, comprising 1600 diverse instances from multiple categories.\footnote{\url{http://www.cril.univ-artois.fr/PB16/bench/PB16-used.tar}} Following the categorization in \cite{pboihs_2021}, we divide PB16 into 42 datasets based on their applications.
    
    \item \texttt{MIPLIB}: A benchmark of 0-1 integer linear programming problems, containing 267 instances of various types, as introduced in \cite{roundingsat_hybrid}.\footnote{\url{https://zenodo.org/record/3870965}}
    
    \item \texttt{CRAFT}: A collection of 1025 crafted combinatorial problems with small integer coefficients, also from \cite{roundingsat_hybrid}.\footnote{\url{https://zenodo.org/record/4036016}}
    
    \item \texttt{Real-world}: A benchmark set containing three application-driven problems: the Minimum-Width Confidence Band Problem (MWCB, 24 instances), the Seating Arrangements Problem (SAP, 21 instances), and the Wireless Sensor Network Optimization Problem (WSNO, 18 instances). All problem descriptions, encodings, and instances are from \cite{lei2021efficient}.\footnote{\label{ft-caisw}\url{https://lcs.ios.ac.cn/\%7ecaisw/Resource/LS-PBO/}}
\end{itemize}

\subsubsection{State-of-the-art Competitors.}

We compare \autopbo with 6 state-of-the-art solvers, including 2 incomplete solver (\ie{} \nupbo and \orasls) and 4 complete solvers. The 4 complete solvers include 2 PB solvers (\ie{} \pboihs and \roundingsat) and 2 MIP solvers (\ie{} \gurobi and \scip):
\begin{itemize}
\item \nupbo \cite{chu2023towards}: The state-of-the-art local search solver for solving PBO.
\item \orasls \cite{iser2023oracle}: a recent oracle-based SLS algorithm for PBO, which improves upon previous pure SLS approaches. 
\item \pboihs \cite{pboihs_2022}: A PBO solver that utilizes the implicit hitting set approach and building upon \roundingsat \cite{roundingsat_v1}.
\item \roundingsat \cite{roundingsat_hybrid}: A PBO solver combining core-guided search with cutting planes reasoning.
\item \gurobi \cite{gurobi2021}: One of the most powerful commercial MIP solvers (Version 12.0.2). The default configuration is used, along with a single thread.
\item \scip \cite{scip}: One of the fastest non-commercial solvers for MIP (Version 8.0.4).
\end{itemize}

\subsubsection{Performance Metrics} 
\begin{table}[t]
\centering
\fontsize{9}{12}\selectfont
\begin{tabular}{l|c|c|c|c}
\hline
\multirow{2}{*}{Benchmark} & \multicolumn{2}{c|}{\structpbo} & \multicolumn{2}{c}{\autopbo} \\
\cline{2-5}
 & \win & \avgscore & \win & \avgscore \\
\hline
Real-world & 17 & 0.9962 & \textbf{29} & \textbf{0.9998} \\
CRAFT & 488 & 0.9472 & \textbf{513} & \textbf{0.9474} \\
MIPLIB & 101 & 0.8450 & \textbf{112} & \textbf{0.8598} \\
PB16 & 697 & 0.8272 & \textbf{775} & \textbf{0.8447} \\
\hline
Total & 1303 & 0.8738 & \textbf{1429} & \textbf{0.8849} \\
\hline
\end{tabular}
\normalsize
\caption{\autopbo vs \structpbo Performance Comparison (By Benchmark)}
\label{tab:improve}
\end{table}

In our experiments, \autopbo first generates optimization strategies on the training set with a 60-second cutoff time. 
Then, each solver performs one run within a given cutoff time (300 seconds) on every instance in the testing set for evaluation. 
We record the cost of the best solution found by solver $S_j$ on instance $I_k$, denoted as $sol_{S_jI_k}$. The cost of the best solution found among all solvers in the same table on instance $I_k$ is denoted as $best_{I_k}$. 

\begin{table*}[ht!]
\centering
\fontsize{9}{12}\selectfont
\setlength{\tabcolsep}{0.5mm}
\begin{tabular}{l|cc|cc|cc|cc|cc|cc|cc}
\hline
  & \multicolumn{2}{c|}{\gurobi} & \multicolumn{2}{c|}{\scip} & \multicolumn{2}{c|}{\pboihs{}-Tuned} & \multicolumn{2}{c|}{\roundingsat{}-Tuned} & \multicolumn{2}{c|}{\orasls{}-Tuned} & \multicolumn{2}{c|}{\nupbo{}-Tuned} & \multicolumn{2}{c}{\autopbo} \\
\cline{2-15}
Benchmark & \win{} & \avgscore{} & \win{} & \avgscore{} & \win{} & \avgscore{} & \win{} & \avgscore{} & \win{} & \avgscore{} & \win{} & \avgscore{} & \win{} & \avgscore{} \\
\hline
Real-world & 3 & 0.5319 & 0 & 0.1417 & 0 & 0.3503 & 0 & 0.0000 & 2 & 0.3322 & 19 & 0.9956 & \textbf{26} & \textbf{0.9977} \\
\hline
CRAFT & \textbf{479} & \textbf{0.9783} & 289 & 0.8337 & 277 & 0.7959 & 302 & 0.8341 & 406 & 0.9620 & 454 & 0.9447 & 460 & 0.9449 \\
\hline
MIPLIB & \textbf{101} & 0.8313 & 33 & 0.5625 & 55 & 0.7051 & 47 & 0.7619 & 53 & 0.7236 & 74 & 0.8259 & 73 & \textbf{0.8367} \\
\hline
PB16 & \textbf{680} & \textbf{0.8461} & 353 & 0.5974 & 513 & 0.7542 & 397 & 0.5297 & 529 & 0.8026 & 600 & 0.8190 & 624 & 0.8204 \\
\hline
Total & \textbf{1263} & \textbf{0.8836} & 675 & 0.6660 & 845 & 0.7555 & 746 & 0.6442 & 990 & 0.8404 & 1147 & 0.8668 & 1183 & 0.8687 \\
\hline
\end{tabular}
\normalsize
\caption{Multi-Solver Performance Comparison (By Benchmark)}
\label{tab:all_benchmark}
\end{table*}

Following previous research on PBO, we measure the performance of each solver using two metrics:
\begin{itemize}
 \item \win: the number of instances where the corresponding $best_{I_k}$ can be obtained by solver $S$ on $B_i$ (\ie{} the number of winning instances).
 \item \avgscore: in our experiments, the competition score of solver $S_j$ on instance $I_k$ is represented by $score_{S_jI_k}=\frac{best_{I_k}+1}{sol_{S_jI_k}+1}$, which measures the gap between $sol_{S_jI_k}$ and $best_{I_k}$.
If solver $S_j$ could not report a solution on instance $I_k$, then $score_{S_jI_k}=0$.  We use \avgscore to denote the average competition score of a solver on a dataset. 
\end{itemize}
For each of the above two metrics, if a solver obtains a larger metric value on a dataset, then the solver exhibits better performance on the dataset. The results highlighted in \textbf{bold} indicate the best performance for the corresponding metric.

\subsection{Results}
\label{sec:exp_results}
\begin{figure}[t]
\centering
\includegraphics[width=\columnwidth]{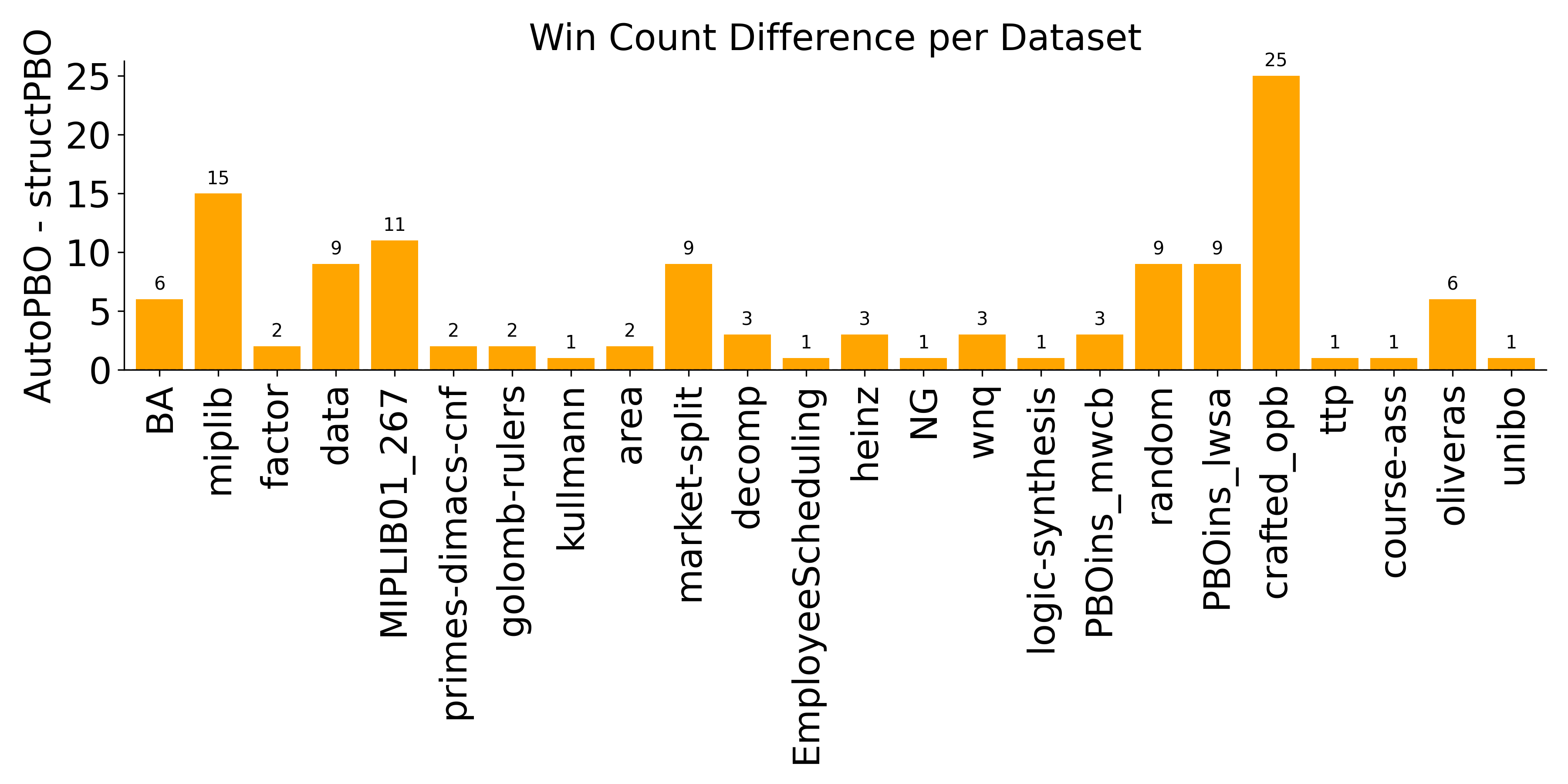} 
\caption{\win differences between \autopbo and \structpbo. Each bar represents \autopbo's \win\ minus \structpbo's \win. Positive values (orange bars) indicate \autopbo's advantage.}
\label{fig:improve_win}
\end{figure}

\begin{figure}[t]
\centering
\includegraphics[width=\columnwidth]{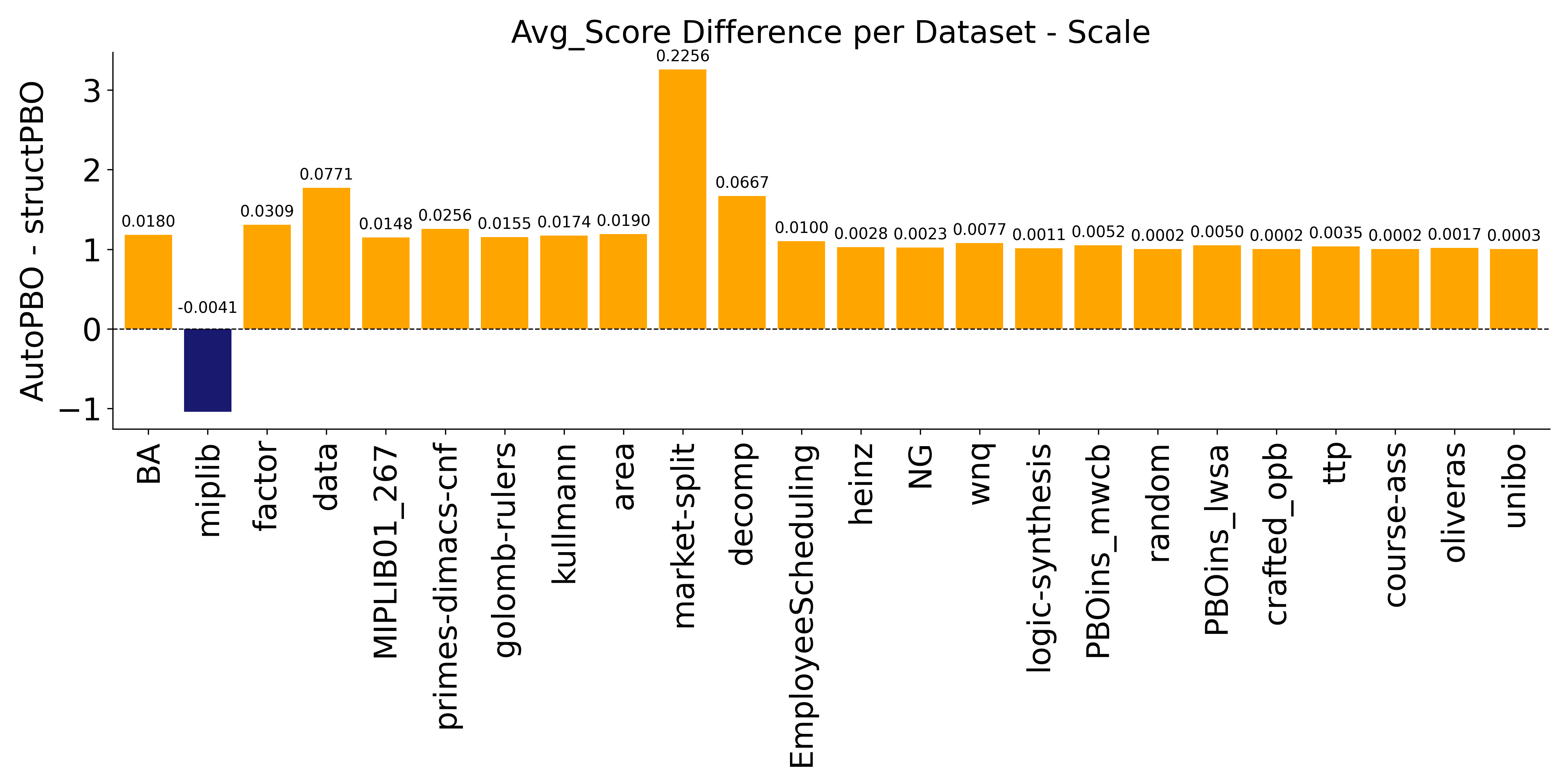} 
\caption{\avgscore\ differences between \autopbo and \structpbo. Each bar represents \autopbo's \avgscore\ minus \structpbo's \avgscore\ (\autopbo $-$ \structpbo). Positive values (orange bars) indicate \autopbo's higher \avgscore, while negative values (dark-blue bars) indicate \structpbo's higher \avgscore (after non-linear scaling for better visibility).}
\label{fig:improve_score}
\end{figure}

\subsubsection{Improvements of Local Search PBO solver}

We first evaluate \autopbo on top of \structpbo across 47 datasets from 4 benchmarks. Figures~\ref{fig:improve_win} and \ref{fig:improve_score} present the datasets with changes in \win and \avgscore respectively along with the magnitude of these changes. \autopbo demonstrates consistent performance gains: improving \win on 24 datasets and \avgscore on 23 datasets, with only a single case of marginal degradation in \avgscore. This consistent non-negative trend confirms that \autopbo not only avoids harming performance but also delivers significant gains on nearly half of the datasets across all benchmarks.

Table~\ref{tab:improve} provides a benchmark-level comparison between \autopbo and \structpbo in terms of both \win and $score$. Across all 4 benchmarks, \autopbo demonstrates consistent improvements, raising the total number of \win and increasing the overall \avgscore. 

These benchmark-level results highlight the generality and robustness of \autopbo.

\subsubsection{Competitive Results of \autopbo}
We conduct a comprehensive comparison between \autopbo and state-of-the-art solvers, as shown in Table \ref{tab:all_benchmark}. To ensure fair comparison, we performed parameter tuning for \pboihs, \roundingsat, \orasls, and \nupbo on each dataset. The tuning scripts and final parameter configurations are documented in Code \& Data Appendix. The experimental results demonstrate that \autopbo outperforms all open-source solvers in both \win and \avgscore. Notably, \autopbo shows competitive performance compared to the commercial solver \gurobi.

At the dataset level (see Appendix A), \autopbo achieves the highest \win on 31 out of 47 datasets and obtains the best \avgscore on 32 datasets. The complete per-dataset comparison reveals that \autopbo consistently delivers robust performance across diverse problem types.

\section{Conclusions and Future Work}
This paper is devoted to develop a novel LLM-powered framework to automatically enhance PBO local search solvers. First, we introduced our structuralized local search PBO solver framework. Furthermore, we configured multiple LLM agents and employed a greedy strategy to generate an optimized and efficient solver. Experimental results demonstrate that AutoPBO can improve the efficiency of local search PBO solver to a very high level.

In the future, we will integrate advanced LLM technologies into more general solvers to enhance their flexibility and performance.  For instance, techniques like retrieval-augmented generation (RAG) could help LLM generating correct modifications. We will also enhance other solvers such as MIP solvers using LLM-based framework. This approach would enable faster modifications and more efficient solver customization.

\bibliography{aaai2026}
\appendix
\onecolumn


\renewcommand{\cite}{\citep} 
\renewcommand{\citet}{\textcite} 


\setlength{\parindent}{10pt} 

\lstset{
    basicstyle          = \footnotesize\ttfamily,
    numbers             = left,
    numberstyle         = \footnotesize,
    xleftmargin         = 2em,
    aboveskip           = 0pt,
    belowskip           = 0pt,
    showstringspaces    = false,
    tabsize             = 2,
    breaklines          = true,
    frame               = single,
    rulecolor           = \color{black},
    framesep            = 4pt
}





\section{Extended Experimental Results}

\subsection{Statistical Analysis of Repeated Experiments}

To ensure the reliability and reproducibility of our experimental results, we conducted multiple independent runs of our experiments to evaluate the stability of solver performance across different conditions.

For each benchmark category (Real-world, CRAFT, MIPLIB, and PB16), we calculated the mean ($\mu$) and standard deviation ($\sigma$) of both win counts and average scores across the experimental runs. The coefficient of variation (CV = $\sigma$/($\mu$) × 100\%) was used as a measure of relative variability to assess the consistency of each solver's performance.

Our statistical analysis reveals that most solvers demonstrate excellent stability across repeated experiments. AutoPBO shows particularly consistent performance with CV values mostly below 5\% for both win counts and scores across all benchmark categories. For instance, in the Real-world benchmark, AutoPBO achieves a win count of 25.3 ± 1.2 (CV = 4.6\%) and a score of 0.997 ± 0.000 (CV = 0.0\%), indicating highly stable performance. Similarly, other solvers like Gurobi, PBO-IHS-Tuned, and RoundingSat-Tuned also exhibit low variability, with most CV values remaining under 3\%.

The low variance observed across multiple experimental runs confirms that our results are not artifacts of specific random initializations or environmental conditions, but rather reflect the genuine algorithmic performance of each solver. This statistical validation strengthens the reliability of our comparative analysis and supports the robustness of our experimental conclusions.

\begin{table*}[t]
\centering
\fontsize{9}{12}\selectfont
\setlength{\tabcolsep}{1mm}
\begin{tabular}{l|cc|cc|cc|cc}
\hline
  & \multicolumn{2}{c|}{Real-world} & \multicolumn{2}{c|}{CRAFT} & \multicolumn{2}{c|}{MIPLIB} & \multicolumn{2}{c}{PB16} \\
\cline{2-9}
Solver & Win($\mu \pm \sigma$) & Score($\mu \pm \sigma$) & Win($\mu \pm \sigma$) & Score($\mu \pm \sigma$) & Win($\mu \pm \sigma$) & Score($\mu \pm \sigma$) & Win($\mu \pm \sigma$) & Score($\mu \pm \sigma$) \\
\hline
\gurobi & 3.3 $\pm$ 0.6 & 0.523 $\pm$ 0.013 & \textbf{479.0 $\pm$ 0.0} & \textbf{0.978 $\pm$ 0.000} & \textbf{102.0 $\pm$ 1.0} & \textbf{0.839 $\pm$ 0.008} & \textbf{680.7 $\pm$ 0.6} & \textbf{0.846 $\pm$ 0.001} \\
\hline
\scip & 0.0 $\pm$ 0.0 & 0.145 $\pm$ 0.005 & 288.7 $\pm$ 0.6 & 0.834 $\pm$ 0.000 & 41.3 $\pm$ 16.2 & 0.561 $\pm$ 0.010 & 358.0 $\pm$ 7.0 & 0.601 $\pm$ 0.007 \\
\hline
\pboihs{}-Tuned & 0.0 $\pm$ 0.0 & 0.351 $\pm$ 0.001 & 276.7 $\pm$ 0.6 & 0.793 $\pm$ 0.005 & 55.0 $\pm$ 0.0 & 0.710 $\pm$ 0.004 & 514.7 $\pm$ 1.5 & 0.753 $\pm$ 0.002 \\
\hline
\roundingsat{}-Tuned & 0.0 $\pm$ 0.0 & 0.000 $\pm$ 0.000 & 301.7 $\pm$ 1.5 & 0.834 $\pm$ 0.000 & 46.7 $\pm$ 0.6 & 0.762 $\pm$ 0.000 & 397.3 $\pm$ 0.6 & 0.529 $\pm$ 0.001 \\
\hline
\orasls{}-Tuned & 2.0 $\pm$ 0.0 & 0.340 $\pm$ 0.013 & 406.0 $\pm$ 0.0 & 0.962 $\pm$ 0.000 & 51.3 $\pm$ 1.5 & 0.723 $\pm$ 0.001 & 529.3 $\pm$ 0.6 & 0.801 $\pm$ 0.001 \\
\hline
\nupbo{}-Tuned & 19.0 $\pm$ 1.0 & 0.996 $\pm$ 0.001 & 454.0 $\pm$ 0.0 & 0.945 $\pm$ 0.000 & 71.7 $\pm$ 2.1 & 0.823 $\pm$ 0.003 & 601.7 $\pm$ 1.5 & 0.818 $\pm$ 0.001 \\
\hline
\autopbo & \textbf{25.3 $\pm$ 1.2} & \textbf{0.997 $\pm$ 0.000} & 460.0 $\pm$ 0.0 & 0.944 $\pm$ 0.001 & 70.7 $\pm$ 2.1 & 0.836 $\pm$ 0.001 & 624.7 $\pm$ 1.2 & 0.819 $\pm$ 0.001 \\
\hline
\end{tabular}
\normalsize
\caption{Solver Stability Analysis Across Experiments}
\label{tab:stability_analysis}
\end{table*}

\subsection{Detailed Experimental Results}
Tables \ref{tab:improve_dataset} and \ref{tab:all_dataset} present the detailed per-dataset evaluation results of \autopbo across all 47 datasets, extending the benchmark-level analysis in Section 5.2. These comprehensive results demonstrate that \autopbo consistently outperforms \structpbo while maintaining competitive performance against state-of-the-art solvers. Specifically, \autopbo achieves superior performance in both \win and \avgscore metrics across diverse problem types, further validating its robust performance.
\begin{table*}[t]
\centering
\fontsize{9}{12}\selectfont
\begin{tabular}{l|c|c|c|c}
\hline
Dataset & \structpbo \win & \autopbo \win & \structpbo \avgscore & \autopbo \avgscore \\
\hline
BA & 5 & \textbf{11 (+6)} & 0.9820 & \textbf{1.0000 (+0.0180)} \\
EmployeeSche & 8 & \textbf{9 (+1)} & 0.8789 & \textbf{0.8889 (+0.0100)} \\
MIPLIB01\_267 & 101 & \textbf{112 (+11)} & 0.8450 & \textbf{0.8598 (+0.0148)} \\
NG & 14 & \textbf{15 (+1)} & 0.9977 & \textbf{1.0000 (+0.0023)} \\
PBOins\_lwsa & 1 & \textbf{10 (+9)} & 0.9949 & \textbf{0.9999 (+0.0050)} \\
PBOins\_mwcb & 7 & \textbf{10 (+3)} & 0.9945 & \textbf{0.9997 (+0.0052)} \\
PBOins\_wsno & \textbf{9} & \textbf{9} & \textbf{1.0000} & \textbf{1.0000} \\
area & 13 & \textbf{15 (+2)} & 0.9810 & \textbf{1.0000 (+0.0190)} \\
areaDelay & \textbf{15} & \textbf{15} & \textbf{1.0000} & \textbf{1.0000} \\
bounded\_golo & \textbf{9} & \textbf{9} & \textbf{0.4444} & \textbf{0.4444} \\
course-ass & 2 & \textbf{3 (+1)} & 0.9998 & \textbf{1.0000 (+0.0002)} \\
crafted\_opb & 488 & \textbf{513 (+25)} & 0.9472 & \textbf{0.9474 (+0.0002)} \\
cudf & \textbf{11} & \textbf{11} & \textbf{0.5455} & \textbf{0.5455} \\
data & 30 & \textbf{39 (+9)} & 0.4452 & \textbf{0.5223 (+0.0771)} \\
decomp & 2 & \textbf{5 (+3)} & 0.9333 & \textbf{1.0000 (+0.0667)} \\
domset & \textbf{8} & \textbf{8} & \textbf{1.0000} & \textbf{1.0000} \\
dt-problems & \textbf{30} & \textbf{30} & \textbf{1.0000} & \textbf{1.0000} \\
factor & 86 & \textbf{88 (+2)} & 0.8756 & \textbf{0.9066 (+0.0310)} \\
fctp & \textbf{16} & \textbf{16} & \textbf{0.1250} & \textbf{0.1250} \\
featureSubsc & \textbf{10} & \textbf{10} & \textbf{1.0000} & \textbf{1.0000} \\
flexray & \textbf{5} & \textbf{5} & \textbf{0.4000} & \textbf{0.4000} \\
frb & \textbf{20} & \textbf{20} & \textbf{1.0000} & \textbf{1.0000} \\
garden & \textbf{4} & \textbf{4} & \textbf{1.0000} & \textbf{1.0000} \\
golomb-ruler & 6 & \textbf{8 (+2)} & 0.6433 & \textbf{0.6589 (+0.0156)} \\
graca & \textbf{10} & \textbf{10} & \textbf{1.0000} & \textbf{1.0000} \\
haplotype & \textbf{4} & \textbf{4} & \textbf{1.0000} & \textbf{1.0000} \\
heinz & 20 & \textbf{23 (+3)} & 0.6494 & \textbf{0.6522 (+0.0028)} \\
kullmann & 3 & \textbf{4 (+1)} & 0.9826 & \textbf{1.0000 (+0.0174)} \\
logic-synthe & 36 & \textbf{37 (+1)} & 0.9989 & \textbf{1.0000 (+0.0011)} \\
market-split & 11 & \textbf{20 (+9)} & 0.3244 & \textbf{0.5500 (+0.2256)} \\
milp & \textbf{17} & \textbf{17} & \textbf{0.7059} & \textbf{0.7059} \\
minlplib & \textbf{49} & \textbf{49} & \textbf{1.0000} & \textbf{1.0000} \\
miplib & 36 & \textbf{51 (+15)} & \textbf{0.7706} & 0.7665 (-0.0041) \\
mps & \textbf{2} & \textbf{2} & \textbf{1.0000} & \textbf{1.0000} \\
oliveras & 57 & \textbf{63 (+6)} & 0.9975 & \textbf{0.9992 (+0.0017)} \\
pbfvmc-formu & \textbf{11} & \textbf{11} & \textbf{1.0000} & \textbf{1.0000} \\
poldner & \textbf{3} & \textbf{3} & \textbf{1.0000} & \textbf{1.0000} \\
primes-dimac & 75 & \textbf{77 (+2)} & 0.8077 & \textbf{0.8332 (+0.0255)} \\
radar & \textbf{6} & \textbf{6} & \textbf{1.0000} & \textbf{1.0000} \\
random & 13 & \textbf{22 (+9)} & 0.7725 & \textbf{0.7727 (+0.0002)} \\
routing & \textbf{8} & \textbf{8} & \textbf{1.0000} & \textbf{1.0000} \\
synthesis-pt & \textbf{5} & \textbf{5} & \textbf{1.0000} & \textbf{1.0000} \\
trarea\_ac & \textbf{5} & \textbf{5} & \textbf{1.0000} & \textbf{1.0000} \\
ttp & 3 & \textbf{4 (+1)} & 0.9965 & \textbf{1.0000 (+0.0035)} \\
unibo & 17 & \textbf{18 (+1)} & 0.3886 & \textbf{0.3889 (+0.0003)} \\
vtxcov & \textbf{8} & \textbf{8} & \textbf{1.0000} & \textbf{1.0000} \\
wnq & 4 & \textbf{7 (+3)} & 0.9895 & \textbf{0.9971 (+0.0076)} \\
\hline
Total & 1303 & \textbf{1429 (+126)} & 0.8738 & \textbf{0.8849 (+0.0110)} \\
\hline
\end{tabular}
\normalsize
\caption{\autopbo vs \structpbo Performance Comparison (By Dataset)}
\label{tab:improve_dataset}
\end{table*}
\begin{table*}[t]
\centering
\fontsize{9}{12}\selectfont
\setlength{\tabcolsep}{0.5mm}
\begin{tabular}{l|cc|cc|cc|cc|cc|cc|cc}
\hline
  & \multicolumn{2}{c|}{\gurobi} & \multicolumn{2}{c|}{\scip} & \multicolumn{2}{c|}{\pboihs{}-Tuned} & \multicolumn{2}{c|}{\roundingsat{}-Tuned} & \multicolumn{2}{c|}{\orasls{}-Tuned} & \multicolumn{2}{c|}{\nupbo{}-Tuned} & \multicolumn{2}{c}{\autopbo} \\
\cline{2-15}
Dataset & \win{} & \avgscore{} & \win{} & \avgscore{} & \win{} & \avgscore{} & \win{} & \avgscore{} & \win{} & \avgscore{} & \win{} & \avgscore{} & \win{} & \avgscore{} \\
\hline
BA & \textbf{13} & \textbf{0.9995} & 0 & 0.2554 & 0 & 0.8295 & 0 & 0.0000 & 1 & 0.8780 & 0 & 0.9676 & 3 & 0.9830 \\
\hline
EmployeeSc & 7 & 0.7579 & 1 & 0.0000 & 2 & 0.5786 & 1 & 0.0000 & 6 & 0.8195 & 8 & 0.8789 & \textbf{9} & \textbf{0.8889} \\
\hline
MIPLIB01\_2 & \textbf{101} & 0.8313 & 33 & 0.5625 & 55 & 0.7051 & 47 & 0.7619 & 53 & 0.7236 & 74 & 0.8259 & 73 & \textbf{0.8367} \\
\hline
NG & \textbf{11} & 0.9466 & 0 & 0.0000 & 0 & 0.8847 & 0 & 0.0000 & 0 & 0.9176 & 7 & \textbf{0.9847} & 7 & \textbf{0.9847} \\
\hline
PBOins\_lws & 0 & 0.0811 & 0 & 0.0000 & 0 & 0.0000 & 0 & 0.0000 & 0 & 0.0000 & 3 & 0.9957 & \textbf{9} & \textbf{0.9997} \\
\hline
PBOins\_mwc & 2 & 0.8082 & 0 & 0.1080 & 0 & 0.6113 & 0 & 0.0000 & 0 & 0.2956 & 9 & 0.9950 & \textbf{10} & \textbf{0.9970} \\
\hline
PBOins\_wsn & 1 & 0.7144 & 0 & 0.3599 & 0 & 0.4305 & 0 & 0.0000 & 2 & 0.7871 & \textbf{7} & \textbf{0.9962} & \textbf{7} & \textbf{0.9962} \\
\hline
area & \textbf{15} & \textbf{1.0000} & 8 & 0.8974 & 10 & 0.9265 & 14 & 0.9892 & 10 & 0.9430 & 13 & 0.9769 & 14 & 0.9949 \\
\hline
areaDelay & \textbf{15} & \textbf{1.0000} & 0 & 0.8691 & 8 & 0.9690 & 14 & 0.9957 & 6 & 0.9683 & \textbf{15} & \textbf{1.0000} & \textbf{15} & \textbf{1.0000} \\
\hline
bounded\_go & 5 & 0.5215 & 3 & 0.2753 & 4 & 0.6140 & 2 & 0.0000 & \textbf{7} & \textbf{0.6610} & 3 & 0.2561 & 3 & 0.3172 \\
\hline
course-ass & \textbf{3} & \textbf{1.0000} & 1 & 0.7863 & 1 & 0.9871 & 2 & 0.9998 & 2 & 0.9976 & 2 & 0.9998 & \textbf{3} & \textbf{1.0000} \\
\hline
crafted\_op & \textbf{479} & \textbf{0.9783} & 289 & 0.8337 & 277 & 0.7959 & 302 & 0.8341 & 406 & 0.9620 & 454 & 0.9447 & 460 & 0.9449 \\
\hline
cudf & 6 & 0.5455 & 6 & 0.5455 & \textbf{10} & \textbf{0.9987} & 6 & 0.5455 & 4 & 0.3636 & 6 & 0.5455 & 6 & 0.5455 \\
\hline
data & \textbf{42} & \textbf{0.6591} & 13 & 0.1445 & 13 & 0.3422 & 18 & 0.3950 & 19 & 0.5079 & 16 & 0.4398 & 17 & 0.4517 \\
\hline
decomp & 1 & 0.7381 & 0 & 0.7788 & 0 & 0.3150 & 0 & 0.0000 & 0 & 0.9355 & 1 & 0.9206 & \textbf{5} & \textbf{1.0000} \\
\hline
domset & 1 & 0.9872 & 0 & 0.9035 & 0 & 0.9774 & 0 & 0.0000 & 0 & 0.9269 & 7 & 0.9993 & \textbf{8} & \textbf{1.0000} \\
\hline
dt-problem & \textbf{30} & \textbf{1.0000} & \textbf{30} & \textbf{1.0000} & \textbf{30} & \textbf{1.0000} & 16 & 0.5333 & 24 & 0.8000 & \textbf{30} & \textbf{1.0000} & \textbf{30} & \textbf{1.0000} \\
\hline
factor & \textbf{96} & \textbf{0.9583} & \textbf{96} & \textbf{0.9583} & \textbf{96} & \textbf{0.9583} & \textbf{96} & \textbf{0.9583} & \textbf{96} & \textbf{0.9583} & \textbf{96} & \textbf{0.9583} & 87 & 0.9006 \\
\hline
fctp & \textbf{16} & \textbf{0.1250} & 15 & 0.1211 & 14 & 0.0000 & 14 & 0.0000 & \textbf{16} & \textbf{0.1250} & \textbf{16} & \textbf{0.1250} & \textbf{16} & \textbf{0.1250} \\
\hline
featureSub & 3 & 0.9869 & 0 & 0.7037 & \textbf{10} & \textbf{1.0000} & \textbf{10} & \textbf{1.0000} & \textbf{10} & \textbf{1.0000} & \textbf{10} & \textbf{1.0000} & \textbf{10} & \textbf{1.0000} \\
\hline
flexray & \textbf{5} & \textbf{0.4000} & \textbf{5} & \textbf{0.4000} & \textbf{5} & \textbf{0.4000} & \textbf{5} & \textbf{0.4000} & 3 & 0.0000 & \textbf{5} & \textbf{0.4000} & \textbf{5} & \textbf{0.4000} \\
\hline
frb & 2 & 0.9966 & 0 & 0.9880 & 0 & 0.9942 & 0 & 0.9808 & 0 & 0.9922 & \textbf{20} & \textbf{1.0000} & \textbf{20} & \textbf{1.0000} \\
\hline
garden & 3 & 0.9057 & 1 & 0.8628 & 3 & 0.7500 & 2 & 0.9256 & 2 & 0.9050 & \textbf{4} & \textbf{1.0000} & \textbf{4} & \textbf{1.0000} \\
\hline
golomb-rul & 6 & 0.6599 & 5 & 0.5486 & \textbf{8} & \textbf{0.8485} & 1 & 0.0000 & 7 & 0.6667 & 4 & 0.5707 & 3 & 0.5740 \\
\hline
graca & \textbf{10} & \textbf{1.0000} & 0 & 0.0098 & 9 & 0.9997 & \textbf{10} & \textbf{1.0000} & \textbf{10} & \textbf{1.0000} & \textbf{10} & \textbf{1.0000} & \textbf{10} & \textbf{1.0000} \\
\hline
haplotype & \textbf{4} & \textbf{1.0000} & 0 & 0.3886 & 3 & 0.9931 & \textbf{4} & \textbf{1.0000} & \textbf{4} & \textbf{1.0000} & \textbf{4} & \textbf{1.0000} & \textbf{4} & \textbf{1.0000} \\
\hline
heinz & 18 & \textbf{0.6515} & 8 & 0.3848 & 15 & 0.6131 & 14 & 0.5739 & 14 & 0.5598 & 18 & 0.6487 & \textbf{19} & 0.6487 \\
\hline
kullmann & 2 & 0.8196 & 0 & 0.3577 & 2 & 0.8122 & 0 & 0.0000 & 1 & 0.6816 & 3 & 0.9817 & \textbf{4} & \textbf{1.0000} \\
\hline
logic-synt & 36 & 0.9986 & 12 & 0.8932 & 35 & 0.9459 & 23 & 0.9250 & 25 & 0.9781 & 36 & 0.9989 & \textbf{37} & \textbf{1.0000} \\
\hline
market-spl & \textbf{20} & \textbf{0.5500} & 11 & 0.1146 & 9 & 0.0130 & 9 & 0.1705 & 12 & 0.3659 & 11 & 0.2544 & 11 & 0.3186 \\
\hline
milp & \textbf{17} & \textbf{0.8235} & 5 & 0.2727 & 9 & 0.5556 & 10 & 0.4745 & 13 & 0.6964 & 7 & 0.5197 & 7 & 0.5389 \\
\hline
minlplib & \textbf{38} & 0.9972 & 9 & 0.9321 & 14 & 0.7229 & 12 & 0.9620 & 10 & 0.9612 & 36 & \textbf{0.9996} & 36 & 0.9994 \\
\hline
miplib & \textbf{52} & \textbf{0.8946} & 16 & 0.3904 & 25 & 0.6048 & 5 & 0.0000 & 27 & 0.7191 & 25 & 0.7028 & 28 & 0.7099 \\
\hline
mps & 1 & \textbf{1.0000} & \textbf{2} & \textbf{1.0000} & 0 & 0.8824 & \textbf{2} & \textbf{1.0000} & \textbf{2} & \textbf{1.0000} & \textbf{2} & \textbf{1.0000} & \textbf{2} & \textbf{1.0000} \\
\hline
oliveras & 47 & 0.7930 & 21 & 0.4640 & 47 & 0.8322 & 47 & 0.8998 & \textbf{70} & \textbf{1.0000} & 42 & 0.9790 & 41 & 0.9800 \\
\hline
pbfvmc-for & 10 & 0.9992 & 1 & 0.5562 & 0 & 0.5031 & 2 & 0.4258 & 3 & 0.5180 & \textbf{11} & \textbf{1.0000} & \textbf{11} & \textbf{1.0000} \\
\hline
poldner & \textbf{3} & \textbf{1.0000} & 2 & 0.9841 & \textbf{3} & \textbf{1.0000} & \textbf{3} & \textbf{1.0000} & \textbf{3} & \textbf{1.0000} & \textbf{3} & \textbf{1.0000} & \textbf{3} & \textbf{1.0000} \\
\hline
primes-dim & \textbf{73} & 0.8205 & 57 & 0.6871 & 68 & 0.7436 & 10 & 0.0000 & 72 & 0.8290 & 70 & 0.8073 & \textbf{73} & \textbf{0.8329} \\
\hline
radar & \textbf{6} & \textbf{1.0000} & 0 & 0.9743 & \textbf{6} & \textbf{1.0000} & 2 & 0.9770 & 3 & 0.9714 & \textbf{6} & \textbf{1.0000} & \textbf{6} & \textbf{1.0000} \\
\hline
random & \textbf{22} & \textbf{0.7727} & 6 & 0.5740 & \textbf{22} & \textbf{0.7727} & \textbf{22} & \textbf{0.7727} & \textbf{22} & \textbf{0.7727} & 13 & 0.7725 & \textbf{22} & \textbf{0.7727} \\
\hline
routing & \textbf{8} & \textbf{1.0000} & 5 & 0.6250 & \textbf{8} & \textbf{1.0000} & \textbf{8} & \textbf{1.0000} & \textbf{8} & \textbf{1.0000} & \textbf{8} & \textbf{1.0000} & \textbf{8} & \textbf{1.0000} \\
\hline
synthesis- & \textbf{5} & \textbf{1.0000} & 3 & 0.9745 & \textbf{5} & \textbf{1.0000} & \textbf{5} & \textbf{1.0000} & 3 & 0.9869 & \textbf{5} & \textbf{1.0000} & \textbf{5} & \textbf{1.0000} \\
\hline
trarea\_ac & \textbf{5} & \textbf{1.0000} & 2 & 0.9822 & 3 & 0.9704 & 0 & 0.0000 & 3 & 0.9636 & \textbf{5} & \textbf{1.0000} & \textbf{5} & \textbf{1.0000} \\
\hline
ttp & 1 & 0.7154 & 1 & 0.7040 & 0 & 0.9296 & 0 & 0.0000 & 1 & 0.9594 & 2 & 0.9894 & \textbf{4} & \textbf{1.0000} \\
\hline
unibo & \textbf{16} & 0.4853 & 8 & 0.1142 & 13 & \textbf{0.4908} & 8 & 0.0000 & 10 & 0.3881 & 8 & 0.2980 & 8 & 0.3161 \\
\hline
vtxcov & 5 & 0.9995 & 0 & 0.9736 & 3 & 0.9986 & 0 & 0.0000 & 0 & 0.9711 & \textbf{8} & \textbf{1.0000} & \textbf{8} & \textbf{1.0000} \\
\hline
wnq & 1 & 0.9719 & 0 & 0.0000 & 0 & 0.9572 & 0 & 0.0000 & 0 & 0.3860 & 4 & 0.9895 & \textbf{7} & \textbf{0.9995} \\
\hline
Total & \textbf{1263} & \textbf{0.8836} & 675 & 0.6660 & 845 & 0.7555 & 746 & 0.6442 & 990 & 0.8404 & 1147 & 0.8668 & 1183 & 0.8687 \\
\hline
\end{tabular}
\normalsize
\caption{Multi-Solver Performance Comparison (By Dataset)}
\label{tab:all_dataset}
\end{table*}

\newpage
\clearpage


\end{document}